# Adaptive Checkpoint Adjoint Method for Gradient Estimation in Neural ODE

**Juntang Zhuang** [1]  **Nicha Dvornek** [1 2]  **Xiaoxiao Li** [1]  **Sekhar Tatikonda** [3]  **Xenophon Papademetris** [1 2 4]
**James Duncan** [1 2 4]

## Abstract

Neural ordinary differential equations (NODEs) have recently attracted increasing attention; however, their empirical performance on benchmark tasks (e.g. image classification) are significantly inferior to discrete-layer models. We demonstrate an explanation for their poorer performance is the inaccuracy of existing gradient estimation methods: the *adjoint* method has numerical errors in reverse-mode integration; the *naive* method directly back-propagates through ODE solvers, but suffers from a redundantly deep computation graph when searching for the optimal stepsize. We propose the *Adaptive Checkpoint Adjoint* (ACA) method: in automatic differentiation, ACA applies a *trajectory checkpoint* strategy which records the forward-mode trajectory as the reverse-mode trajectory to guarantee accuracy; ACA deletes redundant components for shallow computation graphs; and ACA supports *adaptive* solvers. On image classification tasks, compared with the adjoint and naive method, ACA achieves half the error rate in half the training time; NODE trained with ACA outperforms ResNet in both accuracy and test-retest reliability. On time-series modeling, ACA outperforms competing methods. Finally, in an example of the three-body problem, we show NODE with ACA can incorporate physical knowledge to achieve better accuracy. We provide the PyTorch implementation of ACA: https://github.com/juntang-zhuang/torch-ACA.

---

*Equal contribution [1]Department of Biomedical Engineering, Yale University, New Haven, CT USA [2]Department of Radiology & Biomedical Imaging, Yale School of Medicine, New Haven, CT USA [3]Department of Statistics and DataScience, Yale University, New Haven, CT USA [4]Department of Electrical Engineering, Yale University, New Haven, CT USA. Correspondence to: Juntang Zhuang <j.zhuang@yale.edu>, james Duncan <james.duncan@yale.edu>.

## 1. Introduction

Conventional neural networks with discrete layers have achieved great success in various tasks, such as image classification (He et al., 2016), segmentation (Long et al., 2015) and machine translation (Sutskever et al., 2014). However, it's difficult for these discrete-layer networks to model continuous processes. The recently proposed neural ordinary differential equation (NODE) (Chen et al., 2018) views the model as an ordinary differential equation (ODE), whose derivative is parameterized by the network. NODE can be viewed as an initial value problem (IVP), whose initial condition is input to the model. NODE achieves great success in free-form reversible generative models (Grathwohl et al., 2018), time series analysis (Rubanova et al., 2019) and system identification (Quaglino et al., 2019; Ayed et al., 2019). However, the empirical performance of NODE is significantly inferior to discrete-layer models on benchmark classification tasks (error rate: 19% (NODE) vs 5% (ResNet) on CIFAR10) (Dupont et al., 2019; Gholami et al., 2019).

We demonstrate that performance is adversely affected by inaccurate gradient estimation for NODEs using existing methods. NODEs are typically trained with the *adjoint method* (Pontryagin, 1962; Chen et al., 2018), which is memory-efficient but sensitive to numerical errors; because the forward-mode and reverse-mode trajectories are treated as two separate IVPs, they are not accurately equal, causing error in gradient estimation (Gholami et al., 2019). The *naive method* directly back-propagates through ODE solvers; however, it has a redundantly deep computation graph when adaptive solvers search for optimal stepsize (Wanner & Hairer, 1996).

We propose the *adaptive checkpoint adjoint* (ACA) method to accurately estimate the gradient for NODEs. ACA supports adaptive ODE solvers. In automatic differentiation, ACA applies a *trajectory checkpoint* strategy, which stores the forward-mode trajectory with minimal memory; the forward-mode trajectory is used as the reverse-mode trajectory to guarantee numerical accuracy. ACA deletes redundant components during the backward-pass for a shallow computation graph and accurate gradient estimation.

Our contributions can be summarized as:
(1) We theoretically analyze the numerical error with the

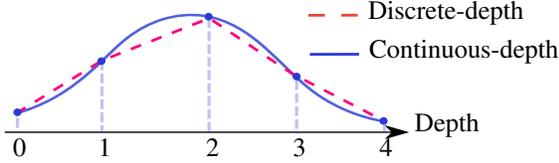

*Figure 1.* From discrete-depth model to continuous depth model.

adjoint and naive methods, and propose ACA to accurately estimate gradients of NODEs.

(2) On image classification tasks, compared with the adjoint and naive methods, ACA achieves twice the speed and half the error rate; furthermore, to our knowledge, ACA is the first to enable NODE with adaptive solvers to outperform ResNet in both accuracy and test-retest reliability. On time series modeling, ACA outperforms other methods.

(3) We show that NODE can incorporate physical knowledge and improve accuracy when trained with ACA. Furthermore, ACA can be applied to general ODEs.

## 2. Preliminaries

### 2.1. Neural Ordinary Differential Equations

NODE views the model as an ordinary differential equation, whose derivative is parameterized by a neural network. NODE can be represented as:

$$\frac{dz(t)}{dt} = f(z(t), t, \theta), \ \ s.t. \ z(0) = x, \ t \in [0, T] \quad (1)$$

where $z(t)$ is the hidden state, $T$ is the end time, and $f$ is the network with parameters $\theta$. $z(0)$ is the initial condition, which equals input $x$. Output of the model is $z(T)$.

We draw the connection between NODE and conventional networks in Fig. 1, where a discrete-depth model takes integer depths, and a continuous-depth model has values at all real-number depths. Compared to discrete-layer models, the feature map evolves smoothly with depth in NODE.

### 2.2. Analytical Form of Gradient for NODE

We formulate the training process of NODE as an optimization problem:

$$argmin_\theta \frac{1}{N} \sum_{i=1}^{N} J(\hat{y}_i, y_i) \quad (2)$$

$$s.t. \ \frac{dz_i(t)}{dt} = f(z_i(t), t, \theta), \ z_i(0) = x_i, \quad (3)$$

$$\hat{y}_i = z_i(T), \ t \in [0, T], \ i = 1, 2, ..N \quad (4)$$

where $J$ is the loss function (e.g. cross-entropy, $L_2$ loss).

We use the Lagrangian Multiplier Method to solve the problem defined in Eq. 4. For simplicity, considering only one

example (can be easily extended to the multiple examples case), the Lagrangian is

$$L = J(z(T), y) + \int_0^T \lambda(t)^\top [\frac{dz(t)}{dt} - f(z(t), t, \theta)] dt \quad (5)$$

**Theorem 2.1** *The gradient derived from Karush-Kuhn-Tucker (KKT) conditions for Eq. 5 is:*

$$\frac{\partial J}{\partial z(T)} + \lambda(T) = 0 \quad (6)$$

$$\frac{d\lambda(t)}{dt} + \left(\frac{\partial f(z(t), t, \theta)}{\partial z(t)}\right)^\top \lambda(t) = 0 \ \ \forall t \in (0, T) \quad (7)$$

$$\frac{dL}{d\theta} = \int_T^0 \lambda(t)^\top \frac{\partial f(z(t), t, \theta)}{\partial \theta} dt \quad (8)$$

Detailed proofs are in Appendix C. $\lambda(t)$ also corresponds to the negative *adjoint variable* in optimal control (Pontryagin, 1962; Chen et al., 2018).

**Summary** The analytical solution can be summarized as:
(1) Solve $z(t)$ in time $0 \to T$.
(2) Determine $\lambda(T)$ with Eq. 6.
(3) Solve $\lambda(t)$ in time $T \to 0$, following Eq. 7 and boundary condition $\lambda(T)$.
(4) Calculate parameter gradient by Eq. 8.

Note that in order to calculate Eq. 8, $\lambda(t)$ and $z(t)$ are required for every $t$. Since $\lambda(t)$ and $z(t)$ are solved in opposite directions, we need to either memorize $z(t)$, or find a method to recover $z(t)$. Note that Eq. 8 is the analytical form, and needs to be numerically calculated in practice.

### 2.3. Numerical Integration

ODE solvers aim to numerically calculate

$$z(T) = z(0) + \int_0^T f(z(t), t, \theta) dt \quad (9)$$

We mainly consider adaptive stepsize solvers. Compared to constant-stepsize solvers, adaptive solvers can estimate error and adaptively control stepsize (Press et al., 1988).

**Notations** We summarize notations here, which are also demonstrated in Fig. 2 and Fig. 3:

- $z_i(t_i)/\overline{z}(\tau_i)$: hidden state in forward/reverse time trajectory at time $t_i/\tau_i$.

- $\Phi_{t_i}^t(z_i)$: the *oracle* solution of the ODE at time $t$, starting from $(t_i, z_i)$. Black dashed curve in Fig. 2 and Fig. 3. $\Phi$ is called the *flow map*.

- $\psi_{h_i}(t_i, z_i)$: the *numerical* solution at time $t_i + h_i$, starting from $(t_i, z_i)$. Blue solid line in Fig. 2.

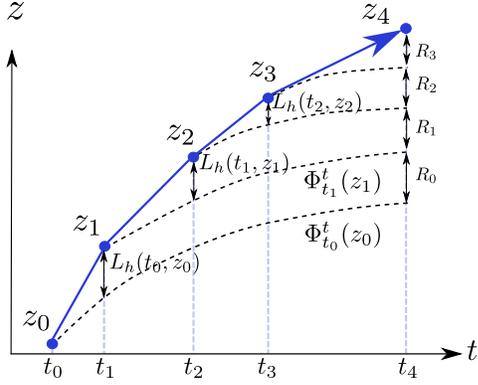

*Figure 2.* Forward-time integration. The global error is the sum of local error $L_h(t_i, z_i)$ propagated to end time. This picture is called *Lady Windermere's Fan* (Wanner & Hairer, 1996). Details of notations summarized in Sec. 2.3

- $L_{h_i}(t_i, z_i)$: local truncation error between numerical approximation and oracle solution, where

$$L_{h_i}(t_i, z_i) = \psi_{h_i}(t_i, z_i) - \Phi_{t_i}^{t_i+h_i}(z_i) \quad (10)$$

- $R_i$: the local error $L_{h_i}(t_i, z_i)$ propagated to end time.

$$R_i = \Phi_{t_{i+1}}^T(z_{i+1}) - \Phi_{t_i}^T(z_i) \quad (11)$$

- $N_f$: number of layers in $f$ in Eq. 1.

- $N_t/N_r$: number of discretized points (outer iterations in Algo. 1) in forward / reverse integration. It varies with input and error tolerance for adaptive solvers.

- $m$: average number of inner iterations in Algo. 1 to find an acceptable stepsize.

---

**Algorithm 1** Numerical Integration

**Input:** data $x$, end time $T$, error tolerance $etol$, initial stepsize $h$

**Initialize:** $z = x, t = 0$, error estimate $\hat{e} = \infty$

**while** $t < T$ **do**

    **while** $\hat{e} > etol$ **do**

        $h \leftarrow h \times decay\_factor(\hat{e})$

        $\hat{e}, \hat{z} = \psi_h(t, z)$

    **end while**

    $t \leftarrow t + h, z \leftarrow \hat{z}$

**end while**

---

The numerical integration algorithm is summarized in Algo. 1 and Fig. 2. The ODE solver progressively advances in time, and adapts the stepsize according to the error estimate. Note that for a given start point $(t_i, z_i)$, the solver might need to execute the inner *while* loop in Algo. 1 many times until the stepsize is small enough, such that the error estimate is below a certain threshold. This process will generate a very redundant deep computation graph, where only the final $h$ is needed.

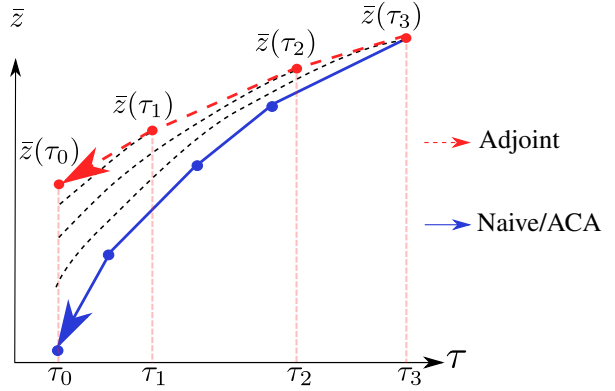

*Figure 3.* Reverse-time integration. Blue curve is the same trajectory as in forward-time integration. Both naive method and ours accurately recover the forward-time trajectory, while adjoint method forgets the forward-time trajectory.

## 3. Methods

In this section we describe different methods to compute the gradient. A summary comparing the methods is given in Table 1 and Fig. 3. We refer readers to the summary part of Sec. 2.2 for the analytical solution; the following methods are different numerical implementations. Note that forward-time integration is the same for these methods.

### 3.1. Summary of Different Methods

**Naive Method: direct back-prop through solver** The simplest way is to treat the numerical ODE solver as a very deep discrete-layer network, and directly back-propagate. We call it the "naive" method. Because all computation graphs (including searching for optimal stepsize) are recorded in the memory for back-prop, the memory cost and depth are $O(N_f \times N_t \times m)$. The computation cost is doubled considering both forward and reverse integration. The memory cost of the naive method can quickly explode, because an accurate solution requires a very small stepsize, and hence a very large $N_t$.

**Adjoint Method: forget forward-time trajectory** To solve the memory issue with the naive method, Chen et al. (2018) proposed the *adjoint method*, originally illustrated by Pontryagin (1962). The adjoint method forgets the forward-time trajectory $z(t)$; instead, it remembers boundary condition $z(T)$ and $\lambda(T)$, then solves $z(t)$ and $\lambda(t)$ in reverse-time $T \to 0$. We use $\bar{z}$ to denote reverse-time solution. Because $\lambda(t)$ and $\overline{z(t)}$ are solved in the same direction, the integration in Eq. 8 only records current values, achieving $O(N_f)$ memory cost. Since the adjoint method needs to solve $\overline{z(t)}$ in reverse-time, it requires extra $O(N_f \times N_r \times m)$ computation, so the total computation cost is $O(N_f \times (N_t + N_r) \times m)$. Note that $\overline{z(t)}$ is not the same as $z(t)$ (as in Fig. 3) due to numerical errors, which will cause error in gradient estimation. We will explain in detail in Sec. 3.2.

**Adaptive Checkpoint Adjoint (ACA): record $z(t)$ with minimal memory**   ACA tries to record $z(t)$ to avoid numerical errors, while also controlling memory cost. ACA supports both adaptive and constant stepsize ODE solvers. It is summarized in Algo. 2, with a detailed version in Appendix A. Note that the forward-pass computation is the same as Algo. 1 for all three methods, so we omit common parts and focus on the unique part.

---

**Algorithm 2** ACA: Record $z(t)$ with Minimal Memory

**Forward-pass:**
  (1) Keep accepted discretization points $\{t_0, ... t_{N_t}\}$
  (2) Keep $z$ values $\{z_0, z_1, ... z_{N_t}\}$ (Not $\psi_{h_i}(t_i, z_i)$)
  (3) Delete local computation graphs to search for optimal stepsize
**Backward-pass:**
  Initialize $\lambda(T)$, $\frac{dL}{d\theta} = 0$
  **For** $N_t$ to 1:
    (1) local forward: $z_{i+1} = \psi(t_i, z_i)$ with stepsize $h_i = t_{i+1} - t_i$
    (2) local backward, update $\lambda$ and $\frac{dL}{\theta}$ according to discretization of Eq. 7 and Eq. 8.
    (3) Delete local computation graphs.

---

During the forward-pass, to save memory, ACA deletes redundant computation graphs to search for the optimal stepsize. Instead, ACA applies the "trajectory checkpoint" strategy, recording the discretization points $t_i$ (equivalently, the accepted stepsize $h_i = t_{i+1} - t_i$) and values $z_i$ (not computation graph $\psi_{h_i}(t_i, z_i)$) at a memory cost $O(N_t)$. Considering $O(N_f)$ memory cost for one evaluation of $\psi$, the total memory cost is $O(N_f + N_t)$.

During the backward-pass, going reverse-time, ACA performs the forward-pass and backward-pass *locally* from $t_i$ to $t_{i+1}$, and updates $\lambda$ and $\frac{dL}{d\theta}$. Computations are evaluated at saved discretization points $\{t_0, ... t_{N_t}\}$, using saved values $\{z_0, ..., z_{N_t}\}$, to guarantee accuracy between forward-time and reverse-time trajectory. We only need to search for optimal stepsize during the forward-pass, with $m$ inner iterations in Algo. 1; during the backward-pass we reuse saved stepsizes, so the total computation cost is $O(N_f \times N_t \times (m+1))$.

### 3.2. Adjoint Method has Numerical Errors

**Numerical Experiments**   Due to memory consideration, the adjoint method forgets forward-time trajectory $z(t)$, and instead solves reverse-time trajectory $\overline{z(\tau)}$ with initial condition $\overline{z(T)} = z(T)$. Thus, $z(t)$ and $\overline{z(\tau)}$ could mismatch due to numerical errors, as demonstrated in Fig. 3. We performed numerical experiments with *ode45* solver under default settings in MATLAB. We experimented with the *van der Pol equation* (Van der Pol, 1960), and a convolutional function with a random $3 \times 3$ kernel. Results are shown in Fig. 4 and 5. These examples validate our argument about the numerical error of the adjoint method.

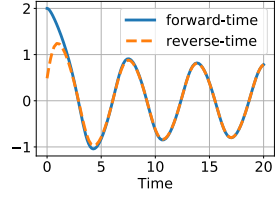

*Figure 4.* Forward time and reverse time trajectory for numerical solution to *van der Pol equation.*

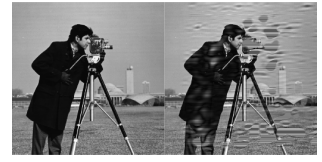

*Figure 5.* Input (left) and reverse-time reconstruction (right) for an ODE defined by a convolution function.

**Analysis of Numerical Errors**   We analyze the numerical errors of the adjoint method. Our results are extensions of Niesen & Hall (2004). We start from the following theorem:

**Theorem 3.1 (Picard-Lindelöf Theorem)** *(Lindelöf, 1894)* Consider the initial value problem (IVP):

$$\frac{dz}{dt} = f(t, z), \;\; z(t = 0) = z_0$$

*Suppose in a region $R = [t_0 - a, t_0 + a] \times [z_0 - b, z_0 + b]$, $f$ is bounded ($||f|| \leq M$), uniformly continuous in $z$ with Lipschitz constant $L$, and continuous in $t$; then there exists a unique solution for the IVP, valid on the region where $a < min\{\frac{b}{M}, \frac{1}{L}\}$.*

The Picard-Lindelöf Theorem states a sufficient condition for existence and uniqueness for an IVP. Okamura (1942) stated a necessary and sufficient condition. Without going deeper, we emphasize that Theorem 3.1 has a validity region, outside this region the theorem may not hold.

It is trivial to check NODE satisfies the above conditions; see the proof in Appendix B. For simplicity, we assume Theorem 3.1 always holds on $t \in [0, T]$. (If $[0, T]$ is outside the region of validity, the adjoint method cannot recover the forward-time trajectory, while the naive method and ACA record the trajectory in memory with "checkpoint".)

Recall $\Phi_{t_i}^t(z_i)$ is the *flow map*, which is the *oracle* solution starting from $(t_i, z_i)$. Define the *variational flow* as:

$$D\Phi_{t_0}^t = \frac{d\Phi_{t_0}^t(z_0)}{dz_0} \tag{12}$$

Consider an ODE solver of order $p$. The local truncation error $L_h(t_i, z_i)$ is of order $O(h^{p+1})$ and can be written as

$$L_h(t_i, z_i) = \psi_h(t_i, z_i) - \Phi_{t_i}^{t_i+h}(z_i) = h^{p+1}l(t_i, z_i) + O(h^{p+2})$$ (13)

where $l$ is some function of $O(1)$. Denote the global error as $G(T)$ at time $T$, then it satisfies:

$$G(T) = z_{N_t} - \Phi_{t_0}^T(z_0) = \sum_{k=0}^{N_t-1} R_k$$ (14)

Eq. 14 is explained by Fig. 2: global error is the sum of all local errors propagated to the end time. $R_k$ is the propagated local error defined by Eq. 11. For simplicity of analysis, we consider constant stepsize solvers with sufficiently small stepsize $h$, and let $N_t = N_r = N$.

**Theorem 3.2** *If the conditions of the Picard-Lindelöf theorem are satisfied, then for an ODE solver of order $p$, the global error at time $T$ is:*

$$G(t_0 \to T) = \sum_{k=0}^{N-1} \left[ h_k^{p+1} D\Phi_{t_k}^T(z_k)l(t_k, z_k) \right] + O(h^{p+1})$$ (15)

*and the error of the reconstructed initial value by the adjoint method is:*

$$
\begin{aligned}
G(t_0 \to T \to t_0) &= G(t_0 \to T) + G(T \to t_0) \\
&= \sum_{k=0}^{N-1} \left[ h_k^{p+1} D\Phi_{t_k}^T(z_k)l(t_k, z_k) + \right. \\
&\left. (-h_k)^{p+1} D\Phi_T^{t_k}(\overline{z_k})\overline{l(t_k, \overline{z_k})} \right] + O(h^{p+1})
\end{aligned}
$$ (16)

*where $G(t_0 \to T \to t_0)$ represents the global error of integration from $t_0$ to $T$, then from $T$ to $t_0$. Terms for reverse-time trajectory are overlined ($\overline{l}, \overline{z}$) to differentiate from forward-time trajectory.*

Proofs are in Appendix B. Eq. 16 can be divided into two parts. $G(t_0 \to T)$ corresponds to forward-time error, as shown in Fig. 2; $G(T \to t_0)$ corresponds to reverse-time error, as shown in Fig. 3. When $h$ is small, assume:

$$\overline{z_k} = z_k + O(h^p)$$ (17)

$$\overline{l(t_k, \overline{z_k})} = l(t_k, z_k) + O(h^p)$$ (18)

$$D\Phi_T^{t_k}(\overline{z_k}) = D\Phi_T^{t_k}(z_k) + O(h^p)$$ (19)

Note that when existence and uniqueness are satisfied, $\Phi$ defines a bijective mapping between $z(t_k)$ and $z(T)$, hence

$$D\Phi_T^{t_k} = (D\Phi_{t_k}^T)^{-1}$$ (20)

Plugging Eq. 17-20 into Eq. 16,

$$G(t_0 \to T \to t_0) = \sum_{k=0}^{N-1} h^{p+1}l(t_k, z_k)e_k + O(h^{p+1})$$ (21)

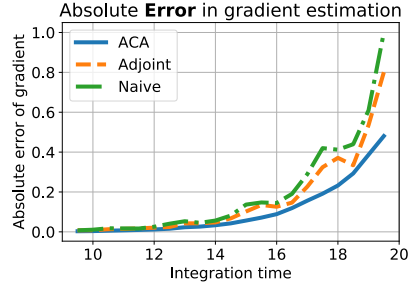

*Figure 6.* Absolute value of error in gradient estimation for different methods. Problem defined by Eq. 27 to Eq. 29.

$$e_k = D\Phi_{t_k}^T(z_k) + (-1)^{p+1}(D\Phi_{t_k}^T(z_k))^{-1}$$ (22)

Reverse accuracy for all $t_0$ requires $e_k = 0$ for all $k$. If $p$ is odd, the two terms in Eq. 22 are of the same sign; thus, $e_k$ cannot be 0. If $p$ is even, $e_k = 0$ requires $D\Phi_{t_k}^T(z_k) = I$, which requires NODE to be an identity function; in this case the model learns nothing. Hence, the adjoint method has numerical errors caused by truncation errors of numerical ODE solvers.

### 3.3. Naive Method has Deep Computation Graph

Note that for each step advance in time, there are on average $m$ steps to find an acceptable stepsize, as in Algo. 1. We give an example below:

$$out_1, h_1, error_1 = \psi(t, h_0, z)$$ (23)
$$out_2, h_2, error_2 = \psi(t, h_1, z)$$ (24)
$$...$$ (25)
$$out_m, h_m, error_m = \psi(t, h_m, z)$$ (26)

Suppose it takes $m$ steps for find an acceptable stepsize such that $error_m < tolerance$. The naive method treats $h_m$ as a recursive function of $h_0$, and back-propagates through all $m$ steps in the computation graph; while ACA takes $h_m$ as a constant, and back-propagates only through the final accepted step (Eq. 26); therefore, the depth of computation graph is $O(N_f \times N_t)$ for ACA, and $O(N_f \times N_t \times m)$ for the naive method. Note that the output of the forward pass is the same for both methods; the backward pass is different.

The very deep computation graph in naive method takes more memory. More importantly, it might cause vanishing or exploding gradient (Pascanu et al., 2013), since there's no special structure such as residual connection to deal with the deep structure: specifically, in Eq. 23 to Eq. 26, only $h_i$ is passed to the next step, and typically in the form $h_{i+1} = h_i/error_i^p$.

### 3.4. ACA Guarantees Reverse-accuracy and has Shallow Computation Graph

Table 1 compares the gradient estimation methods. Adjoint method suffers from numerical error in reverse-mode trajec-tory; naive method suffers from vanishing or exploding gradient caused by deep computation graph ($O(N_f \times N_t \times m)$).

| | Naive | Adjoint | ACA (Ours) |
|---|---|---|---|
| Computation Cost | $O(N_f \times N_t \times m \times 2)$ | $O(N_f \times (N_t + N_r) \times m)$ | $O(N_f \times N_t \times (m+1))$ |
| Memory Consumption | $O(N_f \times N_t \times m)$ | $O(N_f)$ | $O(N_f + N_t)$ |
| Depth of computation graph | $O(N_f \times N_t \times m)$ | $O(N_f \times N_r)$ | $O(N_f \times N_t)$ |
| Reverse accuracy | ✓ | ✗ | ✓ |

*Table 1.* Comparison between different methods to derive parameter gradient. Note $N_r$ is only meaningful for adjoint method. Our method achieves the lowest computation cost, accuracy in reverse-time trajectory, and a shallow computation graph, with medium memory cost.

Compared with the adjoint method, ACA guarantees accuracy of reverse-mode trajectory by recording the forward-mode trajectory. Compared with the naive method, ACA deletes redundant components, hence has a shallower computation graph (ACA v.s. naive: $O(N_f \times N_t)$ v.s. $O(N_f \times N_t \times m)$) and smaller memory cost. Finally, ACA has the lowest computation cost at a medium memory cost.

## 4. Experiments

### 4.1. Toy Example

Consider the following toy problem:

$$\frac{dz(t)}{dt} = kz(t), \;\; z(0) = z_0 \tag{27}$$

$$L(z(T)) = z(T)^2 = z_0^2 exp(2kT) \tag{28}$$

$$\frac{dL}{dz_0} = 2z_0 exp(2kT) \tag{29}$$

We plot the absolute value of error between the analytical solution in Eq. 29 and numerical results from various methods as a function of $T$ in Fig. 6. All numerical methods use the *Dopri5* (Dormand & Prince, 1980) solver with error tolerance $10^{-5}$. ACA consistently outperforms the naive method and adjoint method, which agrees with our analysis in Table 1.

### 4.2. Supervised Learning on Image Classification

**Network Structure** For a fair comparison with state-of-the-art discrete-layer models, we modify a ResNet18 into a NODE18 with the *same* number of parameters. A residual block is defined as:

$$y = x + f(x, \theta) \tag{30}$$

The corresponding ODE-Block is:

$$z(T) = z(0) + \int_0^1 f(z(t), \theta)dt \tag{31}$$

where a residual-block and ODE-Block have the same structure of $f$ (e.g. a sequence of conv-bn-relu layers).

**Comparison of gradient estimation methods for NODE**
We trained the same NODE structure to perform image classification on the CIFAR10 dataset using different gradient estimation methods. The relative and absolute error tolerance are set as 1e-5 for the adjoint and naive method, with *Dopri5* solver implemented by Chen et al. (2018). All methods are trained with SGD optimizer. For each method, we perform 3 runs and record the mean and variance of test accuracy varying with training process. All models are trained for 90 epochs, with initial learning rate of 0.01, and decayed by a factor of 0.1 at epoch 30 and 60. The adjoint method and ACA use a batchsize of 128, while the naive method uses a batchsize of 32 due to its large memory cost.

Test accuracy varying with training epoch is plotted in Fig. 7(a). For the same number of training epochs, ACA ($\sim 5\%$ error rate) outperforms the adjoint and naive method ($\sim 10\%$ error rate) by a large margin.

Test accuracy varying with training time is shown in Fig. 7(b). To train for 90 epochs on a single GTX 1080Ti GPU, ACA takes about 9 hours, while the adjoint method takes about 18 hours, and the naive method takes more than 30 hours. The running time validates our analysis on computation cost in Table 1.

Overall, for the same NODE model, ACA significantly outperforms the adjoint and naive method, with twice the speed and half the error rate.

**Accuracy comparison between NODE and ResNet** We also compare the performance between ResNet and NODE. Note that both models have the same number of parameters.

We trained both models for 10 runs with random initialization using the SGD optimizer. All models are trained for 90 epochs. Results are summarized in Fig. 7(c) and (d). On both CIFAR10 and CIFAR100 datasets, NODE significantly outperforms ResNet when trained with ACA.

We then re-initialized and re-trained for 350 epochs for a fair comparison with ResNet reported by Liu (2017), and summarize the results in Table 2. On image classification tasks, compared to the adjoint method, ACA reduces the error rate of NODE18 from 10% (30%) to 5% (23%) on CIFAR10 (CIFAR100). Furthermore, NODE18 has the same number of parameters as ResNet18, but outperforms deeper networks such as ResNet101 on both datasets.

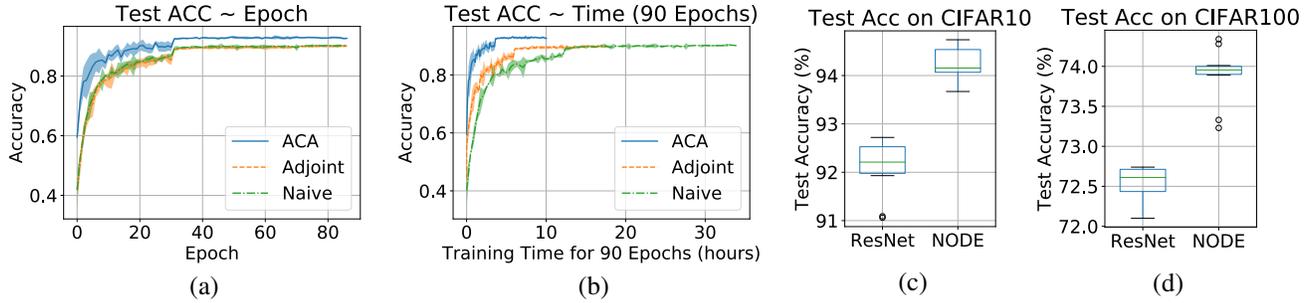

*Figure 7.* From left to right: (a) Test accuracy vs epoch curve on CIFAR10, for NODE18 trained with different methods. (b) Test accuracy vs running time curve on CIFAR10, for NODE18 trained for 90 epochs. (c) Distribution of test accuracy of 10 runs on CIFAR10. NODE18 is trained with ACA. (d) Distribution of test accuracy of 10 runs on CIFAR100. NODE18 is trained with ACA.

| Dataset | NODE18-ACA | | | | | | NODE18 -adjoint | NODE18 -naive | ANODE18 | ResNet | | |
|---|---|---|---|---|---|---|---|---|---|---|---|---|
| | Adaptive Stepsize Solvers | | | Fixed Stepsize Solvers | | | | | | ResNet18 | ResNet50 | ResNet101 |
| | HeunEuler | RK23 | RK45 | Euler | RK2 | RK4 | | | | | | |
| CIFAR10 | **4.85** | 4.92 | 5.29 | 5.52 | 5.27 | 5.24 | 9.8 (*19) | 9.3 | 6.8 | *6.98 | *6.38 | *6.25 |
| CIFAR100 | **22.66** | 24.13 | 23.56 | 24.44 | 24.44 | 24.43 | 30.6 (*37) | 29.4 | 22.7 | *27.08 | *25.73 | *24.84 |

*Table 2.* **Error rate** on test set. NODE18-ACA is trained with HeunEuler solver, and tested with different solvers (including fixed-stepsize and adaptive stepsize solvers of various orders) *without* re-training. Other models are trained and tested with the same method. "*" represents results from the literature (Gholami et al., 2019; He et al., 2016; Liu, 2017), note that our reproduced baseline (adjoint) is better than the literature.

| Dataset | Model | Whole Test Set | | Misclassified Test Data | |
|---|---|---|---|---|---|
| | | ICC1 | ICC1k | ICC1 | ICC1k |
| CIFAR10 | ResNet18 | 0.932-0.935 | 0.992-0.993 | 0.581-0.608 | 0.933-0.939 |
| | NODE18 | **0.943-0.945** | **0.993-0.994** | **0.650-0.675** | **0.949-0.954** |
| CIFAR100 | ResNet18 | 0.759-0.768 | 0.969-0.971 | 0.553-0.571 | 0.925-0.930 |
| | NODE18 | **0.767-0.776** | **0.971-0.972** | **0.570-0.587** | **0.930-0.934** |

*Table 3.* ICC (95% confidence region, $[\mu - 2\sigma, \mu + 2\sigma]$) for ResNet and NODE-ACA among 10 runs with random initialization, tested on CIFAR10 (top) and CIFAR100 (bottom). **Higher** is better.

| Percentage of Training Data | RNN | RNN-GRU | Latent-ODE | | |
|---|---|---|---|---|---|
| | | | adjoint | naive | ACA |
| 10% | *2.45 | *1.97 | 0.47 | *0.36 | **0.31** |
| 20% | *1.71 | *1.42 | 0.44 | *0.30 | **0.27** |
| 50% | *0.79 | *0.75 | 0.40 | *0.29 | **0.26** |

*Table 4.* Test MSE ($\times 10^{-2}$) for irregularly sampled time series data on *Mujoco* dataset. * are reported by Rubanova et al. (2019).

**Robustness to ODE solvers** We implemented adaptive ODE solvers of different orders, as shown in Table 2. HeunEuler, RK23, RK45 are of order 1, 2, 4 respectively, i.e., for each step of $\psi$, $f$ is evaluated 1, 2, 4 times respectively. During inference, using different solvers is equivalent to changing model depth (**without** re-training the network). For discrete-layer models, it would generally cause huge errors (see results in Appendix D); for continuous models, we observe only ~1% increase in error rate. Thus, our method is robust to different solvers.

**Test-retest reliability** Test-retest reliability measures the agreement between multiple raters, and is crucial for clinical practices (Bland & Altman, 1986; Williams et al., 1992).

For machine learning, test-retest reliability quantifies the stability of a model under random initialization and re-training. Intraclass correlation coefficient (ICC) (Weir, 2005) is widely used to quantify test-retest reliability. $ICC$ is between 0 and 1, with higher values for better agreement.

We take the results of 10 runs with independent initialization, as in Fig. 7(c) and (d), and measure $ICC$ with the *psych* package (Revelle, 2017). We report two types of coefficient in Table 3: $ICC1$ (randomly selected judges) and $ICC1k$ (average of raters).

As Table 3 shows, NODE consistently generates higher $ICC$ than ResNet on both datasets. To remove the effect caused by different accuracy, we also measure $ICC$ only on misclassified data points, and NODE produces a significantly higher $ICC$.

**Summary** NODE trained with ACA generates superior performance on benchmark classification tasks. Compared with the adjoint and naive method, ACA is faster and more accurate. Compared with ResNet, to our knowledge, ACA is the first to enable NODE with adaptive solvers to achieve higher accuracy and better test-retest reliability. The better performance comes from two reasons: (1) accurate gradient estimation with ACA, (2) feature maps evolve smoothly with depth (Fig. 1); this property of NODE might be helpful for better generalization (Jin et al., 2019) and optimization (Nesterov, 2005).

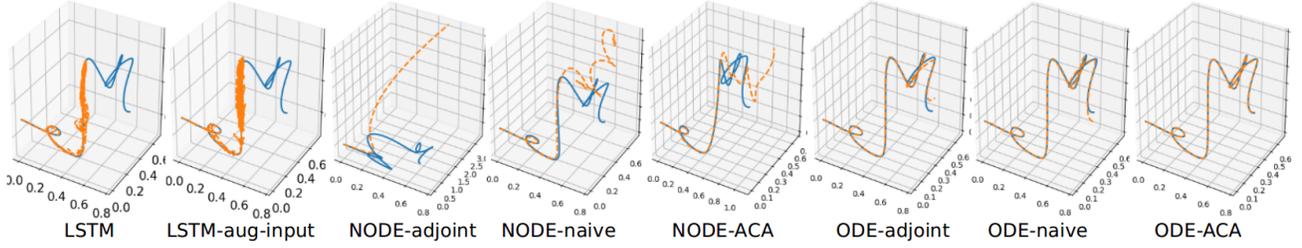

*Figure 8.* Fitted trajectory (orange dashed curve) and ground truth (blue solid curve) for one planet in 3D space. Two trajectories almost overlap in the rightmost figure. Display is adaptively determined in Python for each figure, but ground truth is the same. Time range is [0,1] year for training, and [0,2] years for visualization. Figures are in the same order as Table 5.

| | LSTM | LSTM-aug-input | NODE | | | ODE | | |
| | | | adjoint | naive | ACA | adjoint | naive | ACA |
|---|---|---|---|---|---|---|---|---|
| Test MSE | 0.59+-0.12 | 0.49+-0.06 | 3.47+-0.67 | 0.21+-0.11 | **0.16+-0.06** | 0.0025+-0.0012 | 0.0025+-0.0013 | **0.0007+-0.0005** |

*Table 5.* Results of 3 runs for three-body problem. Training data time range is [0,1] year, MSE is measured on range [0,2] years.

### 4.3. Time-series Model for Irregularly-sampled Data

Standard recurrent neural networks (RNN) have difficulties modelling time series data with non-uniform intervals. The recently proposed latent-ODE model (Rubanova et al., 2019) is a generalization of NODE to time-series models, and can handle arbitrary time gaps.

We validate our method on the *Mujoco* dataset (Tassa et al., 2018) under the same setting as in Rubanova et al. (2019), with the only difference being the gradient estimation method. We report the mean squared error (MSE) of interpolation. As shown in Table 4, ACA consistently outperforms other methods.

### 4.4. Incorporate Physical Knowledge into Models

Differential equations are important tools for modern physics (Sommerfeld, 1949), chemistry (Strogatz, 2018), quantitative biology (Jones et al., 2009), system control (Lee & Markus, 1967) and engineering (Lyapunov, 1992). In practice, for many problems a large training set is *unavailable*, but some physical knowledge is known. It is straightforward to incorporate such knowledge into NODE: set $f$ in the form of physical knowledge.

**Problem definition** We give an example with the three-body problem (Barrow-Green, 1997). Consider three planets (simplified as ideal mass points) interacting with each other, according to Newton's law of motion and Newton's law of universal gravitation (Newton, 1833). The underlying dynamics governing their motion is:

$$\ddot{\mathbf{r}_i} = -\sum_{j \neq i} G m_j \frac{\mathbf{r}_i - \mathbf{r}_j}{|\mathbf{r}_i - \mathbf{r}_j|^3} \qquad (32)$$

where $G$ is the gravitation constant; $\mathbf{r}_i$ is the location for planet $i$, each is of dimension 3; $\ddot{\mathbf{r}_i}$ is the 2nd derivative $w.r.t$ time; $m_i$ is the mass of planet $i$.

We consider the following problem: given observations of trajectory $\mathbf{r}_i(\mathbf{t})$, $t \in [0, T]$, predict future trajectories $\mathbf{r}_i(t)$, $t \in [T, 2T]$, when mass $m_i$ is *unknown*.

**Models** We consider different models:
(1) LSTM with trajectory $\mathbf{r}_i(t)$ as input;
(2) LSTM-aug-input, with augmented input defined as:

$$Aug = \{\mathbf{r}_i, \mathbf{r}_i - \mathbf{r}_j, \frac{\mathbf{r}_i - \mathbf{r}_j}{|\mathbf{r}_i - \mathbf{r}_j|^1}, \frac{\mathbf{r}_i - \mathbf{r}_j}{|\mathbf{r}_i - \mathbf{r}_j|^2}, \frac{\mathbf{r}_i - \mathbf{r}_j}{|\mathbf{r}_i - \mathbf{r}_j|^3}\}, j \neq i \qquad (33)$$

(3) NODE, where $f$ is parameterized as a fully-connected layer using augmented input:

$$\ddot{\mathbf{r}} = FC(Aug) \qquad (34)$$

(4) ODE, with $f$ in the form of Eq. 32. In this case, only 3 parameters, *the mass of planets*, are unknown.

With augmented input, the model knows partial information: the trajectory is related to the distance between planets. The ODE model has full knowledge of the system.

**Results** We simulate a 3-body system with *unequal masses* and *arbitrary initial conditions*, use time range [0,1] year for training, and measure the mean MSE of trajectory on [0,2] years. Results are reported in Fig. 8 and Table 5, with details in Appendix D. We provide videos [1] for better visualization. With no knowledge, the LSTM model fails due to the chaotic nature (Barrow-Green, 1997) of the three-body system and limited training data. With partial knowledge, NODE-ACA outperforms LSTM. With full knowledge, ODE performs the best. ACA outperforms adjoint and naive method, and supports general ODEs.

---

[1] https://www.youtube.com/playlist?list=PL7KkG3n9bER4ODAMzAKzfXIaF0ndUxK-N

## 5. Scopes and limitations

Considering computation burden, this project only investigates explicit general-purpose one-step ODE solvers. There exists rich literature on other solvers including multi-step and implicit solvers (Wanner & Hairer, 1996; Rosenbrock, 1963; Hindmarsh, 1980; Brown et al., 1989). Acceleration methods, such as spectral element methods (Patera, 1984) and parallel-methods (Farhat & Chandesris, 2003), can be used to further improve ACA.

## 6. Related works

**Training of NODE** Quaglino et al. (2019) proposed to use spectral element method to train NODE. However, it requires ground truth for the entire trajectory, and therefore is not suitable for tasks like image classification. Gholami et al. (2019) proposed ANODE to deal with reverse-inaccuracy of the adjoint method, by discretizing the integration range into a *fixed* number of steps. Equivalently, ANODE can be viewed as a fixed-depth discrete-layer network with shared weights. Therefore, ANODE is equivalently using a constant stepsize solver, and loses the ability of error control with adaptive solvers, while ACA supports adaptive solvers. Dupont et al. (2019) proposed to augment NODE to a higher dimension for better performance. However, they are not dealing with gradient estimation, and the empirical performance is still inferior to discrete-layer networks.

**Dynamics and Physics** Many works incorporate physical priors or learn hidden dynamics from data (Ramsay et al., 2007; de Avila Belbute-Peres et al., 2018; Jia et al., 2019; Sienko et al., 2002; Chen & Pock, 2016; Weinan, 2017; Lu, 2017; Sonoda & Murata, 2017). Breen et al. (2019) used deep learning models to solve the three-body problem; however, their methods are limited to *equal masses with zero initial velocities*, while ours do not have these restrictions. Other works try to connect neural networks with dynamical systems (Ruthotto & Haber, 2018; Chang et al., 2018).

**Gradient Checkpointing** The gradient checkpointing (GC) strategy is used to train large networks with limited memory budget (Chen et al., 2016; Gruslys et al., 2016). However, ACA is not a GC version of the naive method: mathematically, naive-GC has the same computation graph depth of $O(N_f \times N_t \times m)$ as naive method, while ACA uses a simplified computation graph of depth $O(N_f \times N_t)$; hence both naive-GC and naive suffer from exploding or vanishing gradient problem, while ACA is more numerically stable; if provided with unlimited memory, naive-GC achieves the same accuracy as naive method, while ACA achieves a higher accuracy.

## 7. Conclusion

We analyzed the inaccuracy of the adjoint and naive method for NODE, and proposed ACA for accurate gradient estimation. We demonstrated NODE trained with ACA is accurate, fast and robust to initialization. Furthermore, NODE can incorporate physical knowledge for better accuracy. We implemented ACA as a package. We hope the good practical performances of ACA, theoretical properties of NODE, and our easy-to-use package, can inspire new ideas.

## Acknowledgement

This research was funded by the National Institutes of Health (NINDS-R01NS035193).

## Appendix A, ACA as an AutoDiff Function in PyTorch Style

**Forward** $(f, T, z_0, tolerance)$:

    $t = 0, z = z_0$

    $state_0 = f.state\_dict()$, cache.save($state_0$)

    Select initial step size $h = h_0$ (adaptively with adaptive step-size solver).

    $time\_points = empty\_list()$

    $z\_values = empty\_list()$

    **While** $t < T$:

        $state = f.state\_dict(),\ \ accept\_step = False$

        **While Not** $accept\_step$:

            $f.load\_state\_dict(state)$

            **with** $grad\_disabled$:

                $z\_new, error\_estimate = \psi(f, z, t, h)$

            **If** $error\_estimate < tolerance$:

                $accept\_step = True$

                $z = z\_new,\ \ t = t + h,$

                $z\_values.append(z),\ \ time\_points.append(t)$

            **else**:

                reduce stepsize $h$ according to $error\_estimate$

                **delete** $error\_estimate$ local computation graph

    cache.save($time\_points,\ \ z\_values$)

    **return** $z$, cache

 

**Backward** $(f, T, tolerance, cache, \frac{\partial J}{\partial z(T)})$:

    **Initialize** $\lambda = -\frac{\partial J}{\partial z(T)},\ \ \frac{\partial L}{\partial \theta} = 0$

    $\{z_0, z_1, z_2, ... z_{N-1}, z_N\} = cache.z\_values$

    $\{t_0, t_1, t_2, ... t_{N-1}, t_N\} = cache.time\_points$

    **For** $t_i$ in $\{t_N, t_{N-1}, ..., t_1, t_0\}$ :

        **Local forward** $\hat{z}_i = \psi(f, z_{i-1}, t_{i-1}, h_i = t_i - t_{i-1})$

        **Local backward**

            $\frac{\partial L}{\partial \theta} \leftarrow \frac{\partial L}{\partial \theta} - \lambda^\top \frac{\partial \hat{z}_i}{\partial \theta}$

           $\lambda \leftarrow \lambda^\top \frac{\partial \hat{z}_i}{\partial z_{i-1}}$

        **delete** local computation graph

    **return** $\frac{\partial L}{\partial \theta},\ \ \lambda$

## Appendix B, Proof of Theorem 3.2

We refer readers to Fig. 2 in the main paper for a graphical interpretation, and Sec. 2.3 for a list of notations.

**Lemma 7.1** *Suppose $f$ is composed of a finite number of ReLU activations and linear transforms,*

$$f(t, z) = Linear_1 \circ ReLU \circ Linear_2 \circ ... \circ Linear_N(z)$$

*if the spectral norm of linear transform is bounded, then the IVP defined above has a unique solution on a bounded region.*

*Proof:* $f$ does not depend on $t$ explicitly, hence is continuous in $t$. ReLU (and other activation functions such as sigmoid, tanh, ...) is uniformly continuous; a linear transform $Wz$ is also uniformly continuous if the spectral norm of $W$ is bounded. From Picard-Lindelöf Theorem, the IVP has a unique solution on a bounded region.

**Flow map** Denote $\Phi_{t_0}^T(z_0)$ as the *oracle* solution to the IVP at time $T$, with the initial condition $(t_0, z_0)$. Then $\Phi_{t_0}^T(z_0)$ satisfies:

$$\Phi_{t_2}^{t_3} \circ \Phi_{t_1}^{t_2} = \Phi_{t_1}^{t_3} \tag{35}$$

$$\frac{d}{dt}\Phi_{t_0}^t(z_0) = f(t, \Phi_{t_0}^t(z_0)) \tag{36}$$

$$\Phi_{t_0}^t(z_0) = z_0 + \int_{t_0}^t f(s, \Phi_{t_0}^s(z_0))ds \tag{37}$$

**Variational flow** The derivative $w.r.t$ initial condition is called the *variational flow*, denoted as $D\Phi_{t_0}^t$, then it satisfies:

$$D\Phi_{t_0}^t(z_0) = \frac{d\Phi_{t_0}^t(z_0)}{dz_0}, \;\; D\Phi_{t_0}^{t_0} = I \tag{38}$$

$$D\Phi_{t_0}^{t_0+h} = I + O(h), \; if \; h \; is \; small. \tag{39}$$

From Eq. 35 and 39, using the chain rule, we have:

$$D\Phi_{t_0}^t(z_0) = \frac{d\Phi_{t_0}^t(z_0)}{dz_0} = \frac{d\Phi_{t_0+h}^t(\Phi_{t_0}^{t_0+h}(z_0))}{d\Phi_{t_0}^{t_0+h}(z_0)}\frac{d\Phi_{t_0}^{t_0+h}(z_0)}{dz_0} = D\Phi_{t_0+h}^t + O(h) \tag{40}$$

**Local truncation error** Denote the step function of a one-step ODE solver as $\psi_h(t, z)$, with step-size $h$ starting from $(t, z)$. Denote the local truncation error as:

$$L_h(t, z) = \psi_h(t, z) - \Phi_t^{t+h}(z) \tag{41}$$

For a solver of order $p$, the error is of order $O(h^{p+1})$, and can be written as

$$L_h(t, z) = h^{p+1}l(t, z) + O(h^{p+2}) \tag{42}$$

where $l$ is some function of order $O(1)$.

**Global error** Denote the global error as $G(T)$ at time $T$, then it satisfies:

$$G(T) = z_N - \Phi_{t_0}^T(z_0) = \sum_{k=0}^{N-1} R_k \tag{43}$$

where

$$R_k = \Phi_{t_{k+1}}^T(z_{k+1}) - \Phi_{t_k}^T(z_k) \tag{44}$$

$$= \Phi_{t_{k+1}}^T\big(\Phi_{t_k}^{t_{k+1}}(z_k) + L_{h_k}(t_k, z_k)\big) - \Phi_{t_{k+1}}^T(\Phi_{t_k}^{t_{k+1}}(z_k)) \tag{45}$$

**Lemma 7.2 (Approximation of $R_k$)**

$$R_k = D\Phi_{t_{k+1}}^T\big(\Phi_{t_k}^{t_{k+1}}(z_k)\big)L_{h_k}(t_k, z_k) + O(h_k^{2p+2}) \tag{46}$$

Lemma 7.2 can be viewed as a Taylor expansion of Eq. 45, with detailed proof in (Niesen & Hall, 2004).

**Lemma 7.3** *If $L_h(t, y) = O(h^{p+1})$, then $G_h(T) = O(h^p)$*

Proof for Lemma 7.3 is in (Niesen & Hall, 2004).

Plug Eq. 42 and Eq. 40 into Eq. 46, we have

$$R_k = \big[D\Phi_{t_k}^T(z_k) + O(h_k)\big]L_{h_k}(t_k, z_k) + O(h_k^{2p+2}) \tag{47}$$

$$= \big[D\Phi_{t_k}^T(z_k) + O(h_k)\big]\big[h_k^{p+1}l(t_k, z_k) + O(h_k^{p+2})\big] + O(h_k^{2p+2}) \tag{48}$$

$$= h_k^{p+1}D\Phi_{t_k}^T(z_k)l(t_k, z_k) + O(h_k^{p+2}) \tag{49}$$

Plug Eq. 49 into Eq. 43, then we have:

$$G(T) = \sum_{k=0}^{N-1} R_k = \sum_{k=0}^{N-1}\big[h_k^{p+1}D\Phi_{t_k}^T(z_k)l(t_k, z_k) + O(h_k^{p+2})\big] \tag{50}$$

$$= \sum_{k=0}^{N-1}\big[h_k^{p+1}D\Phi_{t_k}^T(z_k)l(t_k, z_k)\big] + O(h_{max}^{p+1}) \tag{51}$$

**Global error of the adjoint method**   If we solve an IVP forward-in-time from $t = 0$ to $T$, then take $z(T)$ as the initial condition, and solve it backward-in-time from $T$ to 0, the numerical error can be written as:

$$G(t_0 \to T \to t_0) = \sum_{k=0}^{N_t-1}\big[h_k^{p+1}D\Phi_{t_k}^T(z_k)l(t_k, z_k)\big] + \sum_{J=0}^{N_r-1}\big[(-h_j)^{p+1}D\Phi_T^{\tau_j}(\overline{z_j})\overline{l(\tau_j, \overline{z_j})}\big] + O(h_{max}^{p+1}) \tag{52}$$

$$= G(t_0 \to T) + G(T \to t_0) + O(h^{p+1}) \tag{53}$$

where $G(t_0 \to T)$ represents the numerical error of forward-in-time ($t_0$ to $T$) solution (discretized at step $k$, denoted as $z_k$); and $G(T \to t_0)$ denotes the numerical error of reverse-in-time solution ($T$ to 0) (discretized at step $j$, denoted as $\overline{z_j}$). $G(t_0 \to T \to t_0)$ represents the error in reconstructed initial condition by the adjoint method. Note that generally $z$ does not overlap with $\overline{z}$. The local error of forward-in-time and reverse-in-time numerical integration is represented as $l$ and $\overline{l}$ respectively.

Although going backward is equivalent to a negative stepsize, which might cause the second term to have different signs compared to the first term in Eq. 52, we demonstrate that generally their sum cannot cancel.

For the ease of analysis, we assume the forward and reverse-in-time calculation are discretized at the same grid points, with a sufficiently small constant stepsize (For a variable-stepsize solver, we can modify it to a constant-stepsize solver, whose stepsize is the minimal step in variable-stepsize solver. With this modification, the constant stepsize solver should be *no worse than* adaptive stepsize solver). Then Eq. 52 can be written as:

$$G(t_0 \to T \to t_0) = \sum_{k=0}^{N-1}\big[h_k^{p+1}D\Phi_{t_k}^T(z_k)l(t_k, z_k)\big] + \sum_{k=0}^{N-1}\big[(-h_k)^{p+1}D\Phi_T^{t_k}(\overline{z_k})\overline{l(t_k, \overline{z_k})}\big] + O(h_{max}^{p+1}) \tag{54}$$

$$= \sum_{k=0}^{N-1}\big[h_k^{p+1}D\Phi_{t_k}^T(z_k)l(t_k, z_k) + (-h_k)^{p+1}D\Phi_T^{t_k}(\overline{z_k})\overline{l(t_k, \overline{z_k})}\big] + O(h^{p+1}) \tag{55}$$

If the stepsize is sufficiently small, we can assume

$$z_k = \overline{z_k} + O(h) \tag{56}$$

$$D\Phi_T^{t_k}(z_k) = D\Phi_T^{t_k}(\overline{z_k}) + O(h) \tag{57}$$

$$l(t_k, z_k) = \overline{l(t_k, \overline{z_k})} + O(h) \tag{58}$$

Assume the existence and uniqueness conditions are satisfied on $t \in [0, T]$, so $\Phi_{t_0}^T$ defines a homeomorphism, hence:

$$D\Phi_T^{t_k} = (D\Phi_{t_k}^T)^{-1} \tag{59}$$

Plug Eq. 56 to Eq. 59 into Eq. 55, we have

$$G(t_0 \to T \to t_0) = \sum_{k=0}^{N-1} h^{p+1} l(t_k, z_k) e_k + O(h^{p+1}) \tag{60}$$

$$e_k = D\Phi_{t_k}^T(z_k) + (-1)^{p+1}(D\Phi_{t_k}^T(z_k))^{-1} \tag{61}$$

Reverse accuracy for all $t_0$ requires $e_k = 0$ for all $k$. If $p$ is odd, then the two terms in $e_k$ are the same sign, and thus cannot cancel to 0; if $p$ is even, then $e_k = 0$ requires $D\Phi_{t_k}^T(z_k) = D\Phi_{t_k}^T(z_k)^{-1} = I$, which is generally not satisfied with a trained network (otherwise the network is an identity function with a constant bias).

In short, solving an IVP from $t_0$ to $T$ with $z(0) = z_0$, then taking $z(T)$ as initial condition and solving it from $T$ to $t_0$ and getting $\overline{z(0)}$, generally $z(0) \neq \overline{z(0)}$ because of numerical errors.

## Appendix C. Proof of Theorem 2.1

In this section we derive the gradient in NODE from an optimization perspective.

**Notations**  With the same notations as in the main paper, we use $z(t)$ to denote hidden states $z$ at time $t$. Denote parameters of $f$ as $\theta$, and input as $x$, target as $y$, and predicted output as $\hat{y}$. Denote the loss as $J(\hat{y}, y)$. Denote the integration time as 0 to $T$.

**Problem setup**  The continuous model is defined to follow an ODE:

$$\frac{dz(t)}{dt} = f(z(t), t, \theta), \ \ s.t. \ z(0) = x \tag{62}$$

We assume $f$ is differentiable almost everywhere, since $f$ is represented by a neural network in our case. The forward pass is defined as:

$$\hat{y} = z(T) = z(0) + \int_0^T f(z(t), t, \theta) dt \tag{63}$$

The loss function is defined as:

$$J(\hat{y}, y) = J(z(T), y) \tag{64}$$

We formulate the training process as an optimization problem:

$$\underset{\theta}{\operatorname{argmin}} \frac{1}{N} \sum_{i=1}^{N} J(\hat{y}_i, y_i) \ \ s.t. \ \frac{dz_i(t)}{dt} = f(z_i(t), t, \theta), \ \ z_i(0) = x_i, \ \ i = 1, 2, ... N \tag{65}$$

For simplicity, Eq. 65 only considers one ODE block. In the case of multiple blocks, $z(T)$ is the input to the next ODE block. As long as we can derive $\frac{dLoss}{d\theta}$ and $\frac{dLoss}{dz(0)}$ when $\frac{dLoss}{dz(T)}$ is given, the same analysis here can be applied to the case with a chain of ODE blocks.

**Lagrangian Multiplier Method**   We use the Lagrangian Multiplier Method to solve the problem defined in Eq. 65. For simplicity, only consider one example (can be easily extended to the multiple examples case), then the Lagrangian is

$$L = J(z(T), y) + \int_0^T \lambda(t)^\top [\frac{dz(t)}{dt} - f(z(t), t, \theta)]dt \tag{66}$$

Karush-Kuhn-Tucker (KKT) conditions are necessary conditions for a solution to be optimal. In the following sections we start from the KKT conditions and derive our results.

**Derivative w.r.t.** $\lambda$   At optimal point, we have $\frac{\delta L}{\delta \lambda} = 0$. Note that $\lambda$ is a function of $t$, and we derive the derivative from calculus of variation.

Consider a continuous and differentiable perturbation $\overline{\lambda(t)}$ on $\lambda(t)$, and a scalar $\epsilon$, $L$ now becomes a function of $\epsilon$,

$$L(\epsilon) = J\big(z(0) + \int_0^T f(z(t), t, \theta), y\big) + \int_0^T \big(\lambda(t) + \epsilon\overline{\lambda(t)}\big)^\top [\frac{dz(t)}{dt} - f(z(t), t, \theta)]dt \tag{67}$$

It's easy to check the conditions for Leibniz integral rule, and we can switch integral and differentiation, thus:

$$\frac{dL}{d\epsilon} = \int_0^T \overline{\lambda(t)}^\top [\frac{dz(t)}{dt} - f(z(t), t, \theta)]dt \tag{68}$$

At optimal $\lambda(t)$, $\frac{dL}{d\epsilon}|_{\epsilon=0} = 0$ for all continuous differentiable $\overline{\lambda(t)}$.

Therefore,

$$\frac{dz(t)}{dt} - f(z(t), t, \theta) = 0, \ \ \forall t \in (0, T) \tag{69}$$

**Derivative w.r.t** $z$   Consider perturbation $\overline{z(t)}$ on $z(t)$, with scale $\epsilon$. With similar analysis:

$$L(\epsilon) = J(z(T) + \epsilon\overline{z(T)}, y) + \int_0^T \lambda(t)^\top [\frac{dz(t) + \epsilon\overline{z(t)}}{dt} - f(z(t) + \epsilon\overline{z(t)}, t, \theta)]dt \tag{70}$$

Take derivative w.r.t $\epsilon$, it's easy to check conditions for Leibniz integral rule are satisfied, when $f$ and $\overline{z(t)}$ are Lipschitz continuous differentiable functions:

(1) $f(z(t), t, \theta)$ is a Lebesgue-integrable function of $\theta$ for each $z(t) \in \mathbf{R}^d$, since we use a neural network to represent $f$, which is continuous and differentiable almost everywhere.

(2) for almost all $\theta$, $\frac{\partial f(z(t), t, \theta)}{\partial z(t)}$ exists for almost all $x \in \mathbf{R}^d$.

(3) $\frac{\partial f(z(t), t, \theta)}{\partial z(t)}$ is bounded by $g(\theta)$ for all $z(t)$ for almost all $\theta$.

Then we calculate $\frac{dL(\epsilon)}{d}$, note that we can switch integral and derivative:

$$\frac{dL}{d\epsilon}\big|_{\epsilon=0} = \frac{\partial J}{\partial z(T)}^{\top}\overline{z(T)} + \frac{d}{d\epsilon}\int_0^T \lambda(t)^{\top}\big[\frac{dz(t) + \epsilon\overline{z(t)}}{dt} - f(z(t) + \epsilon\overline{z(t)}, t, \theta)\big]dt \tag{71}$$

$$= \frac{\partial J}{\partial z(T)}^{\top}\overline{z(T)} + \int_0^T \lambda(t)^{\top}\big[\frac{d\overline{z(t)}}{dt} - \frac{\partial f(z(t), t, \theta)}{\partial z(t)}\overline{z(t)}\big]dt \tag{72}$$

$$= \frac{\partial J}{\partial z(T)}^{\top}\overline{z(T)} + \int_0^T \big[\lambda(t)^{\top}\frac{d\overline{z(t)}}{dt} + \frac{d\lambda(t)}{dt}^{\top}\overline{z(t)} - \frac{d\lambda(t)}{dt}^{\top}\overline{z(t)} - \lambda(t)^{\top}\frac{\partial f(z(t), t, \theta)}{\partial z(t)}\overline{z(t)}\big]dt \tag{73}$$

$$= \frac{\partial J}{\partial z(T)}^{\top}\overline{z(T)} + \lambda(t)^{\top}\overline{z(t)}\big|_0^T - \int_0^T \overline{z(t)}^{\top}\big[\frac{d\lambda(t)}{dt} + \frac{\partial f(z(t), t, \theta)}{\partial z(t)}^{\top}\lambda(t)\big]dt \tag{74}$$

$$= \frac{\partial J}{\partial z(t)}^{\top}\overline{z(T)} + \lambda(t)^{\top}\overline{z(T)} - \lambda(0)^T\overline{z(0)} - \int_0^T \overline{z(t)}^{\top}\big[\frac{d\lambda(t)}{dt} + \frac{\partial f(z(t), t, \theta)}{\partial z(t)}^{\top}\lambda(t)\big]dt \tag{75}$$

$$= \big(\frac{\partial J}{\partial z(T)} + \lambda(T)\big)^{\top}\overline{z(T)} - \lambda(0)^T\overline{z(0)} - \int_0^T \overline{z(t)}^{\top}\big[\frac{d\lambda(t)}{dt} + \frac{\partial f(z(t), t, \theta)}{\partial z(t)}^{\top}\lambda(t)\big]dt \tag{76}$$

Since the initial condition $z(0) = x$ is given, perturbation $\overline{z(0)}$ at $t = 0$ is 0, then we have:

$$\frac{dL}{d\epsilon}\big|_{\epsilon=0} = \big(\frac{\partial J}{\partial z(T)} + \lambda(T)\big)^{\top}\overline{z(T)} - \int_0^T \overline{z(t)}^{\top}\big[\frac{d\lambda(t)}{dt} + \frac{\partial f(z(t), t, \theta)}{\partial z(t)}^{\top}\lambda(t)\big]dt = 0 \tag{77}$$

for any $\overline{z(t)}$ $s.t.$ $\overline{z(0)} = 0$ and $\overline{z(t)}$ is differentiable.

The solution is:

$$\frac{\partial J}{\partial z(T)} + \lambda(T) = 0 \tag{78}$$

$$\frac{d\lambda(t)}{dt} + \frac{\partial f(z(t), t, \theta)}{\partial z(t)}^{\top}\lambda(t) = 0 \quad \forall t \in (0, T) \tag{79}$$

**Derivative w.r.t $\theta$** From Eq. 66,

$$\frac{dL}{d\theta} = -\int_0^T \lambda(t)^{\top}\frac{\partial f(z(t), t, \theta)}{\partial \theta}dt \tag{80}$$

To sum up, we first solve the ODE forward-in-time with Eq. 62, then determine the boundary condition by Eq. 78, then solve the ODE backward with Eq. 79, and finally calculate the gradient with Eq. 80. In fact $\lambda$ corresponds to the negative *adjoint variable*.

## Appendix D, Experimental Details

### 1. Experiments with van der Pol Equation

For experiments with the *van der Pol equation*, the ODE is defined as:

$$\frac{dy_1}{dt} = y_2 \tag{81}$$

$$\frac{dy_2}{dt} = (0.15 - y_1^2) \times y_2 - y_1 \tag{82}$$

with initial condition $y_1(0) = 2, y_2(0) = 0$. Experiments are performed in MATLAB with *ode45* solver under default settings. Results are shown as follow:

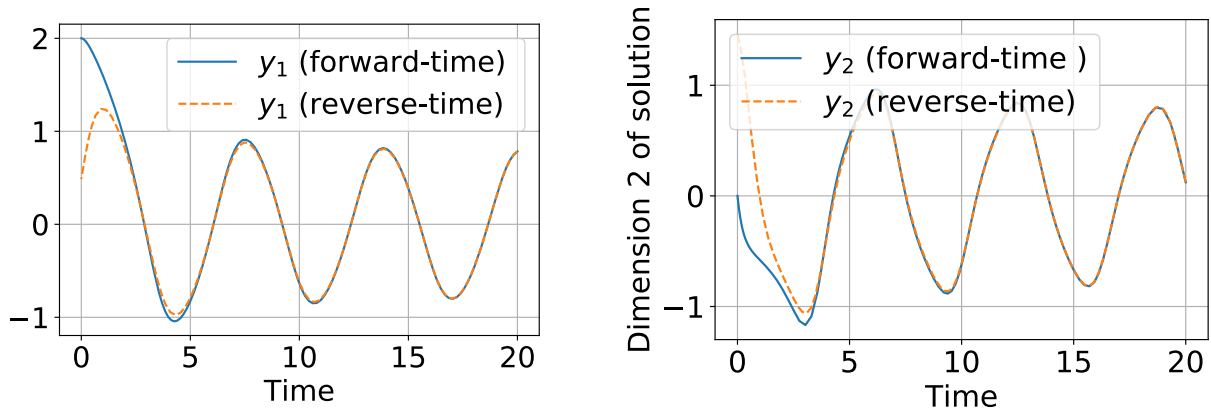

*Figure 9.* Results on simulation with van der Pol equation.

### 2. Experiments on supervised image classification

**Experimental settings**  All experiments were performed with PyTorch 0.4.1 on a single GTX-1080Ti GPU. To generate Fig. 7 (a) and (b) in the main paper, we trained a NODE18 model for 90 epochs, with the initial learning rate 0.01 and decayed by a factor of 0.1 at epoch 30 and 60. Training images were augmented with random crop and horizontal flip. For ACA, we used *HeunEuler* solver for training, with $rtol = 10^{-2}$ and $atol = 10^{-2}$; for the adjoint and naive method, we used the solver implemented by (Chen et al., 2018) (`https://github.com/rtqichen/torchdiffeq`), with a *Dopri5* solver, setting $rtol = 10^{-5}$ and $atol = 10^{-5}$; we tried larger error tolerance ($10^{-2}$) for the adjoint method, but it led to divergence during training. Batchsize is set as 128 for ACA and the adjoint method, and 32 for the naive method.

To generate Fig. 7(c) and (d), and Table 3 in the main paper, we trained NODE18-ACA and ResNet with the following settings: initial learning rate is 0.1, decayed by 0.1 at epoch 30 and 60, total training epoch is 90; batchsize is set as 128.

To generate Table 2 in the main paper, we trained NODE18-ACA for 350 epochs, with initial learning rate 0.1, and decayed by 0.1 at epoch 150 and 250.

**Extra experiments on impact of model depth**  We performed experiments on CIFAR10 dataset to measure the influence of model depth. We trained models using the same settings as described above, and tested with different solvers *without* re-training.

During test, constant stepsize solvers using different stepsizes are equivalent to different model depths, for example, with a stepsize of 0.2, the model depth is 5 times deeper than with a stepsize of 1.0 ($\frac{1.0}{0.2} = 5$); higher-order solvers evaluates the function more times than low-order solvers, for example, using the same stepsize, RK4 evaluates 4 times wile RK2 evaluates twice at each step. Adaptive stepsize solvers evaluate the function using a finer grid for smaller error tolerance, hence a deeper computation graph.

|  | Constant Stepsize Solvers | | | | Adaptive Stepsize Solvers | | | |
|---|---|---|---|---|---|---|---|---|
| stepsize | 1.0 | 0.5 | 0.2 | 0.1 | rtol / atol | 1e-1 | 1e-2 | 1e-3 |
| Euler | **+0.0** | +1.14 | +3.62 | +4.84 | HeunEuler | +5.32 | +6.33 | +6.42 |
| RK2 | +7.69 | +6.43 | +6.38 | +6.43 | RK23 | +6.03 | +6.31 | +6.44 |
| RK4 | +5.69 | +6.31 | +6.30 | +6.47 | RK45 | +6.30 | +6.47 | +6.46 |

*Table 6.* **Increase in error rate** of a ResNet18 (equivalently, NODE using 1-step Euler method with integration time [0,1]) when tested with different solvers. When trained and tested with the same method, the error rate is 8.47%, the increase in error rate is 0 as bold fonted. When tested with different solvers **without** re-training, the increase in error rate is reported, with a **smaller** difference represents better robustness.

|  | Constant Stepsize Solvers | | | | Adaptive Stepsize Solvers | | | |
|---|---|---|---|---|---|---|---|---|
| stepsize | 1.0 | 0.5 | 0.2 | 0.1 | rtol / atol | 1e-1 | 1e-2 | 1e-3 |
| Euler | +8.31 | +1.57 | +0.67 | +0.57 | HeunEuler | +1.29 | **+0.0** | +0.18 |
| RK2 | +6.61 | +0.57 | +0.42 | +0.39 | RK23 | +0.46 | +0.07 | +0.40 |
| RK4 | +1.09 | +0.48 | +0.39 | +0.37 | RK45 | +1.75 | +0.44 | +0.16 |

*Table 7.* **Increase in error rate** of a NODE18 when trained using HeunEuler with $rtol = atol = 10^{-2}$ and tested with different solvers. When trained and tested with the same method, the error rate is 4.85%, the increase in error rate is 0 as bold fonted. When tested with different solvers **without** re-training, the increase in error rate is reported, with a **smaller** difference represents better robustness.

To sum up, depth of the computation graph is determined by both the *stepsize* and *order* for constant stepsize solvers, and determined by *error tolerance* and *order* for adaptive stepsize solvers.

We performed experiments on a ResNet18; equivalently, ResNet18 is NODE18 using one-step Euler solver, with integration time $[0, 1]$. We also experimented with a NODE18, trained with HeunEuler solver with $rtol = etol = 10^{-2}$. Results for their performance using different solvers *without* re-training are summarized in Table. 6 and Table. 7.

NODE generally achieves a lower error rate than ResNet (4.85% v.s 8.47% when trained and tested using the same method). Ignoring the absolute value of error rate, to measure the robustness to solvers, on the *same* model, we focus on the *increase* in test error rate using different methods compared to using the same method as training.

For ResNet, when tested with different solvers, most results have a $\sim 7\%$ increase in error rate; for NODE when trained with HeunEuler with $rtol = 10^{-2}$, and tested with different solvers, most results have a $\sim 1\%$ increase in error rate. This results show that training as ResNet is sensitive to model depth during test; while training as NODE with adaptive solvers are robust to different solvers (hence different model depth) during test.

## 3. Experiments on time-series modeling with irregularly sampled data

We performed experiments using the official implementation by (Rubanova et al., 2019) `https://github.com/YuliaRubanova/latent_ode`. All models are trained for 300 epochs on the *Mujoco* dataset provided by (Rubanova et al., 2019). For the ease of visualization, we plot the test MSE curve for epochs 0-100.

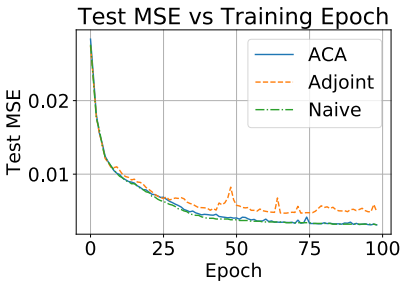

*Figure 10.* 10% training data

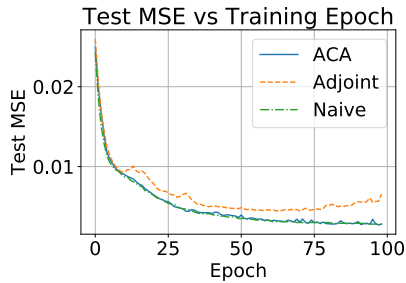

*Figure 11.* 20% training data

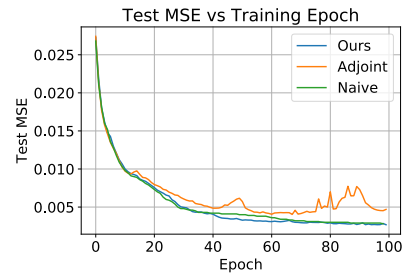

*Figure 12.* 50% training data

## 4. Experiments on the three-body problem

We summarize the training details in the following table. All models are trained with Adam optimizer. Learning rate schedule is:

$$lr = InitialLR \times decay^{epoch} \tag{83}$$

For all LSTM models, initial learning rate is 0.01 with a $decay$ of 0.999, trained for 5,000 epochs; all NODEs are trained with initial learning rate 0.1 for 100 epochs, with $decay$ 0.99. We set a much larger epoch and smaller learning rate for LSTM, because we found in practice the training of LSTM is much harder to converge. For adjoint and naive method, we use $Dopri5$ solver by (Chen et al., 2018) with $rtol = atol = 10^{-5}$; for ACA, we implemented $Dopri5$ solver with $rtol = atol = 10^{-5}$.

We simulate a 3-body system with unequal mass and arbitrary initial condition, use time range [0,1] year for training, and measure the mean MSE of trajectory on [0,2] years. Training data has 1,000 equally sampled points, cut into sequences of 20 points as input to LSTM models. During inference, 1 initial point is fed to NODE and ODE, and first 10 points are fed to LSTM. Results are shown in figures below and videos in the supplementary material, with numerical measures in the main paper.

| | LSTM | LSTM-aug-input | NODE | | | ODE | | |
|---|---|---|---|---|---|---|---|---|
| | | | adjoint | naive | ACA | adjoint | naive | ACA |
| Epoch | 5,000 | 5,000 | 100 | 100 | 100 | 100 | 100 | 100 |
| InitialLR | 0.01 | 0.01 | 0.1 | 0.1 | 0.1 | 0.1 | 0.1 | 0.1 |
| decay | 0.999 | 0.999 | 0.99 | 0.99 | 0.99 | 0.99 | 0.99 | 0.99 |

*Table 8.* Training details for the three-body problem

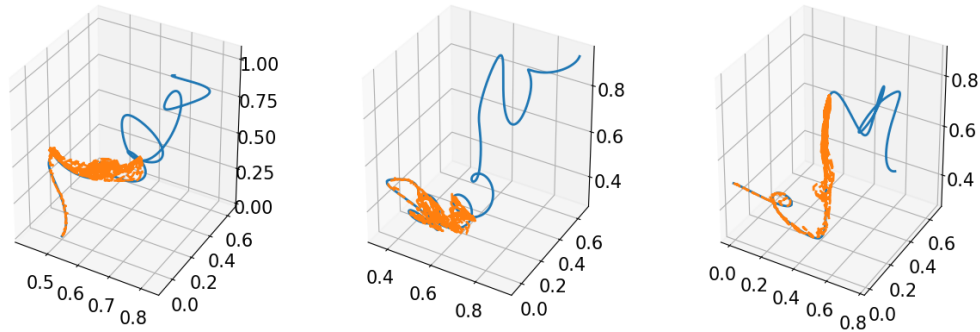

*Figure 13.* Fitted trajectory (orange dashed) and ground truth (blue solid) in 3D space, from left to right are results for 3 planets. Results for LSTM model.

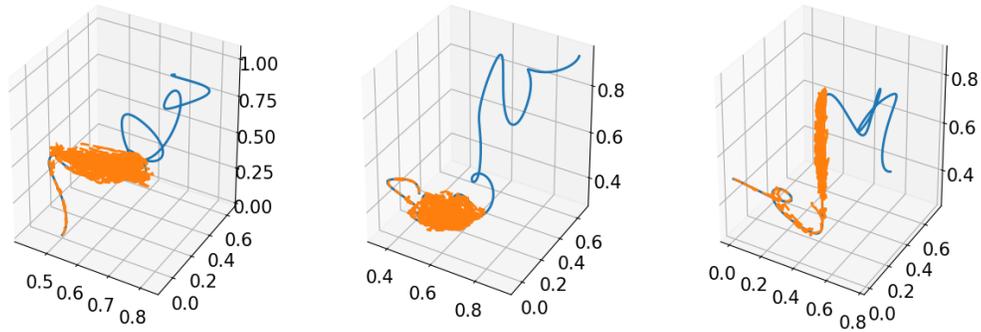

*Figure 14.* Fitted trajectory (orange dashed) and ground truth (blue solid) in 3D space, from left to right are results for 3 planets. Results for LSTM-aug-input.

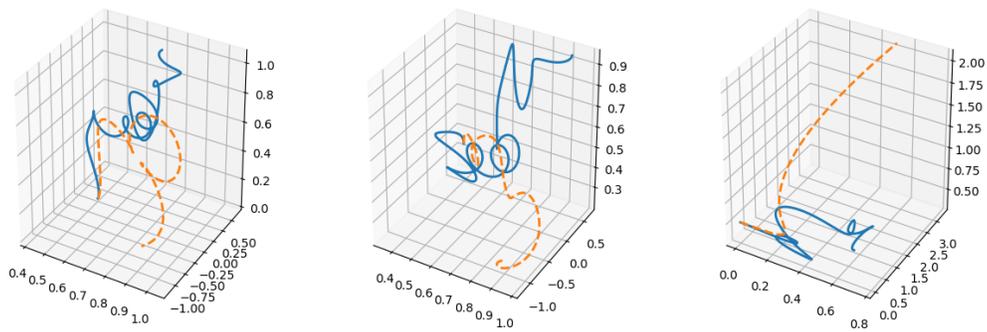

*Figure 15.* Fitted trajectory (orange dashed) and ground truth (blue solid) in 3D space, from left to right are results for 3 planets. Results for NODE-adjoint.

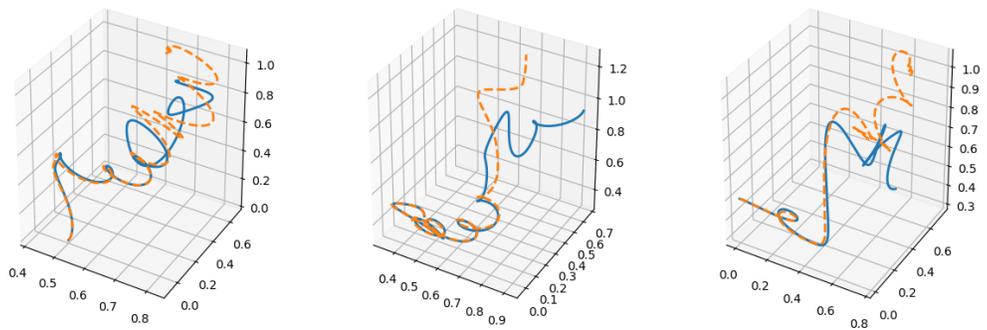

*Figure 16.* Fitted trajectory (orange dashed) and ground truth (blue solid) in 3D space, from left to right are results for 3 planets. Results for NODE-naive.

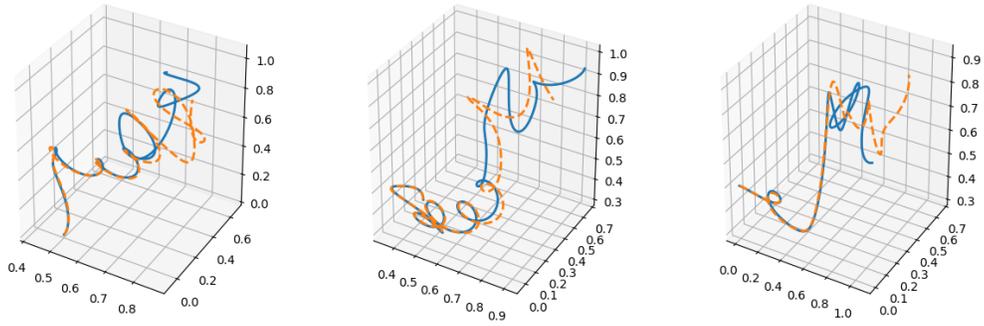

*Figure 17.* Fitted trajectory (orange dashed) and ground truth (blue solid) in 3D space, from left to right are results for 3 planets. Results for NODE-ACA.

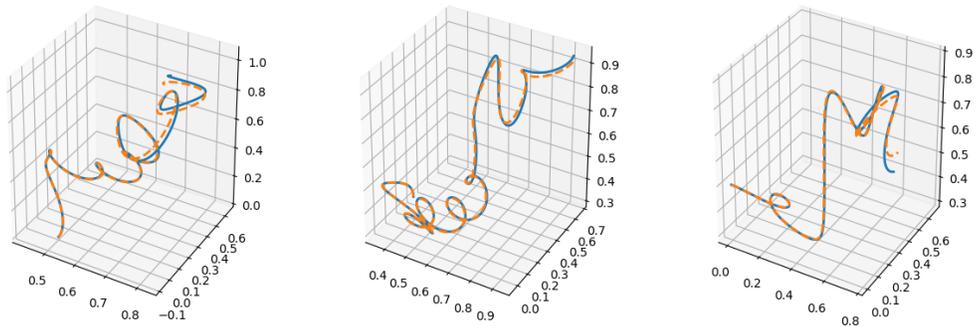

*Figure 18.* Fitted trajectory (orange dashed) and ground truth (blue solid) in 3D space, from left to right are results for 3 planets. Results for ODE-adjoint.

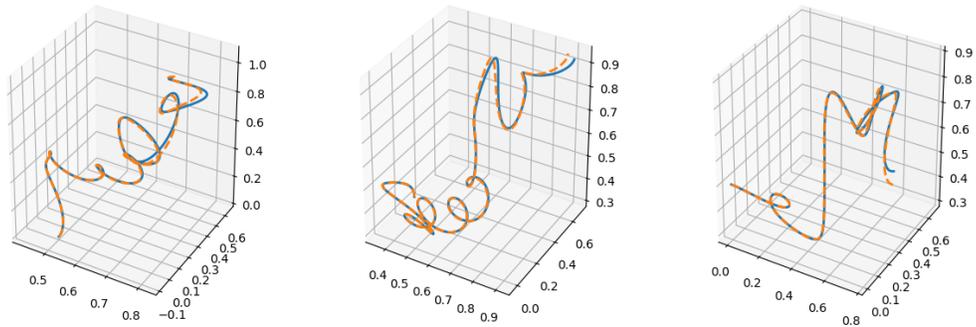

*Figure 19.* Fitted trajectory (orange dashed) and ground truth (blue solid) in 3D space, from left to right are results for 3 planets. Results for ODE-naive.

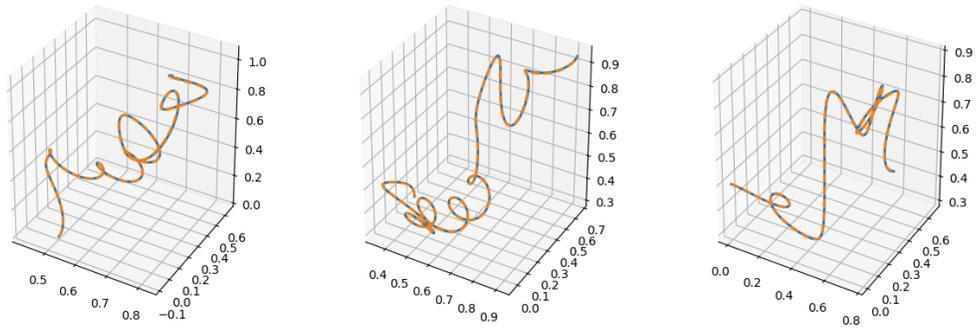

*Figure 20.* Fitted trajectory (orange dashed) and ground truth (blue solid) in 3D space, from left to right are results for 3 planets. Results for ODE-ACA.

## More results of ODE-ACA with different initial conditions

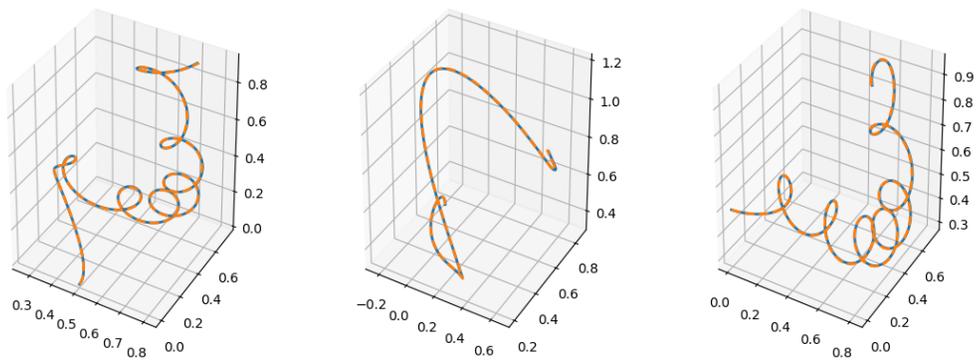

*Figure 21.* Fitted trajectory (orange dashed) and ground truth (blue solid) in 3D space, from left to right are results for 3 planets. Results for ODE-ACA.

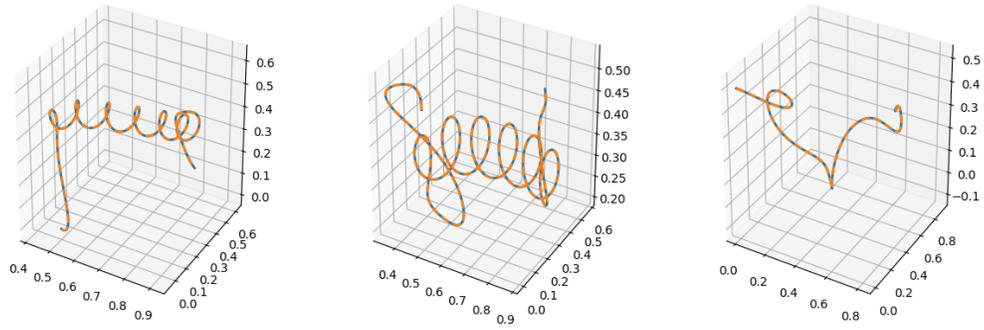

*Figure 22.* Fitted trajectory (orange dashed) and ground truth (blue solid) in 3D space, from left to right are results for 3 planets. Results for ODE-ACA.

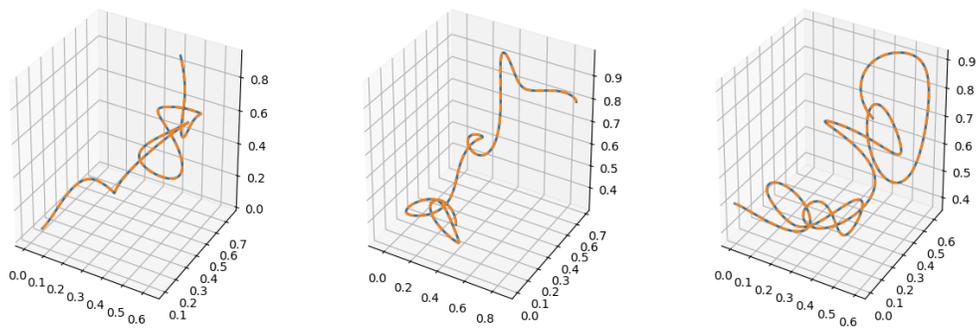

*Figure 23.* Fitted trajectory (orange dashed) and ground truth (blue solid) in 3D space, from left to right are results for 3 planets. Results for ODE-ACA.